\documentclass[runningheads]{llncs}

\usepackage[mobile]{eccv}

\usepackage{eccvabbrv}

\usepackage{graphicx}
\usepackage{booktabs}

\usepackage{amsmath}
\usepackage{booktabs}
\usepackage{multirow}
\usepackage{xcolor}
\usepackage{array}
\usepackage[table]{xcolor}
\usepackage{enumitem}
\usepackage{amsmath}

\usepackage{wrapfig}
\usepackage{makecell}
\usepackage{wrapfig}

\usepackage[accsupp]{axessibility}  % Improves PDF readability for those with disabilities.

\usepackage{hyperref}

\usepackage{orcidlink}

\begin{document}

\title{Occ-VLM: Occupancy Grounded Vision Language Model for Indoor Scene Understanding} 

\titlerunning{Occ-VLM}

\author{Jianing Li \and
Zhou Fang \and
Yijiang Liu \and
Li Du}

\authorrunning{J. Li et al.}

\institute{School of Electronic Science and Engineering, Nanjing University}

\maketitle

\begin{abstract}
Recently, vision-language models (VLMs) have made significant progress in 3D scene understanding, driving advances in applications such as embodied intelligence and robotic vision. However, existing approaches typically either rely directly on explicit 3D inputs (e.g., point clouds or RGB-D sequences), or introduce an additional 3D geometry encoder to derive 3D-aware visual tokens from 2D images. Such designs structurally decouple 3D geometric perception from the rich 2D semantics learned via vision-language pre-training, hindering the development of a unified 3D vision-language representation. In this work, we propose Occ-VLM, a novel framework for 3D scene understanding that operates purely on posed RGB images and employs a single 2D vision encoder. Specifically, Occ-VLM reconstructs 3D scene occupancy as an auxiliary geometric prior, which is utilized to spatially associate foreground 2D tokens with 3D space. These tokens are then decoded by a Large Language Model (LLM) for unified scene understanding. Extensive experiments demonstrate that Occ-VLM achieves both accurate geometric perception and robust vision-language reasoning: it attains state-of-the-art performance on multi-view occupancy prediction, while performing on par with 3D-input VLMs on 3D Visual Question Answering (VQA) and 3D dense captioning benchmarks. Code is available at \url{https://github.com/NorthSummer/Occ-VLM.git}.
\end{abstract}

\section{Introduction}
\label{sec:intro}

Vision-Language Models (VLMs)\cite{alayrac2022flamingo ,liu2023visual, liu2024improved, li2023blip, bai2023qwen, wang2024qwen2} have achieved remarkable progress on multimodal understanding tasks. Mainstream VLMs are primarily designed for 2D visual understanding. Given images or videos as input, a pretrained 2D vision encoder extracts visual tokens, which are then projected into the language space, fused with text tokens, and finally fed into a Large Language Model (LLM) for reasoning and generation. In this way, VLMs can follow human instructions and answer questions grounded in image and video content, thereby achieving a unified cross-modal understanding of visual information.

\begin{figure}[t]
    \centering
    \includegraphics[width=\linewidth]{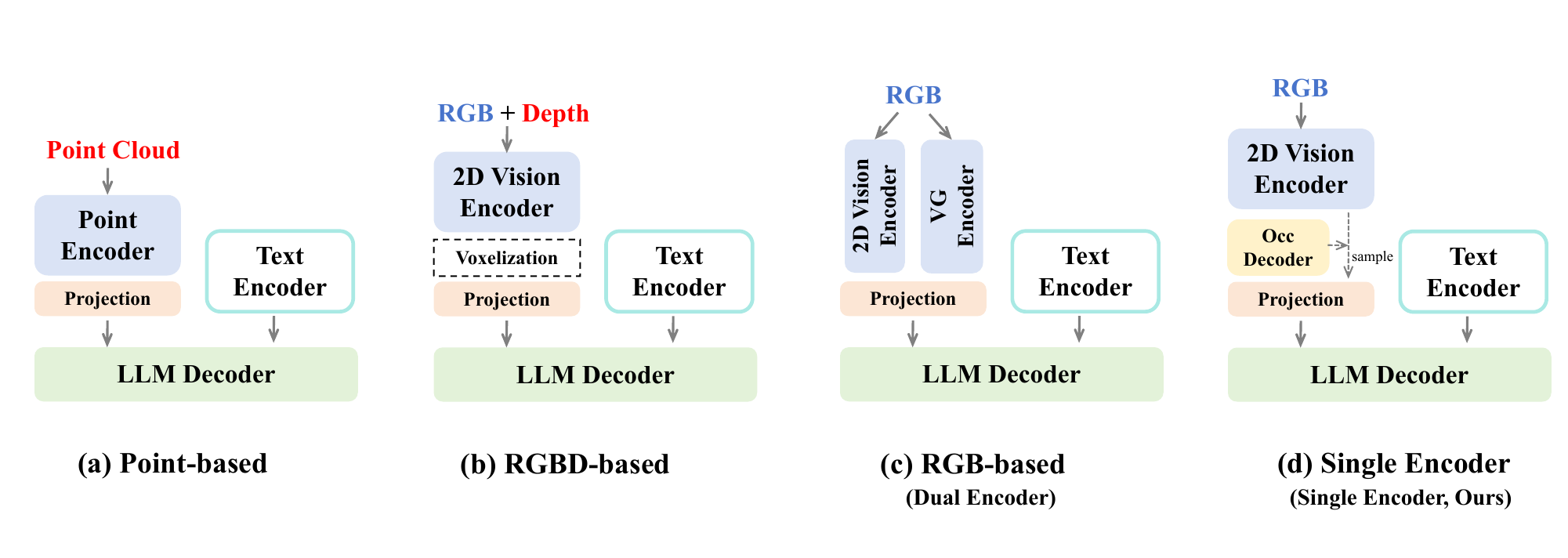}
    \caption{ Comparison of 3D VLM architectures. (a) Point-based and (b) RGBD-based methods both rely on explicit 3D data inputs. (c) Dual-encoder methods enable RGB-only scene understanding but necessitate an additional geometric encoder. (d) Our proposed Occ-VLM achieves RGB-only scene understanding using a single 2D encoder.}
    \label{fig:cover}
\end{figure}

Building upon the success of VLMs in 2D visual understanding, an increasing number of studies have explored how to extend their capabilities to 3D scene understanding (Fig. \ref{fig:cover}). Early works primarily employed 3D data such as point clouds as input\cite{guo2023point, qi2024gpt4point, xu2024pointllm}. These methods typically use a point-based 3D encoder to extract 3D tokens and align them with text tokens. However, due to the unstructured nature of point clouds, such methods often require complex alignment processes at both the object and scene levels. Subsequently, research shifted towards utilizing RGB-D data\cite{zhu2024llava, zheng2025video, qi2025gpt4scene}. These methods exploit explicit depth information to unproject 2D tokens into 3D space through techniques like 3D positional encoding and voxelization, significantly simplifying the cross-modal alignment. Nevertheless, due to their strict reliance on depth sensors, their application scenarios remain limited; when only RGB inputs are available, their performance degrades noticeably.

To overcome this reliance on explicit 3D data, recent research has begun to focus on RGB-only 3D VLM architectures. These methods\cite{wu2025spatial, zheng2025learning} typically take RGB videos as input and introduce a frozen pre-trained visual geometry encoder (e.g., a VGGT\cite{wang2025vggt}) alongside task-specific decoders. This dual-encoder architecture reconstructs the scene geometry from monocular inputs to extract 3D features, which are then fused with 2D tokens. Although this paradigm improves geometric perception accuracy without requiring explicit 3D inputs, it introduces high architectural complexity and computational overhead. This raises an important question: Can we design a VLM capable of comprehensive 3D scene understanding using purely RGB inputs and a single 2D encoder?

To address this question, we propose Occ-VLM. Given posed RGB images, we first employ an Occ adapter to lift the 2D encoder features into 3D occupancy grids, utilizing an occupancy decoder to predict 3D semantic occupancy and explicitly recover scene geometry. Based on this, we further introduce an occupancy-aware 3D token representation. Specifically, this module samples occupied regions in 3D space and projects them onto the 2D feature planes to sample corresponding features, thereby generating spatially aware foreground 3D tokens to be fed into the LLM for spatial reasoning. In contrast to prior works that treat 3D perception as an isolated task, we view it as a bidirectional 2D–3D bridge. Our central insight is that while occupancy prediction facilitates geometric reconstruction, the rich semantics encoded in the 2D encoder can, in turn, significantly enhance 3D scene recovery. Extensive experiments support this perspective: on the multi-view 3D semantic occupancy prediction task, Occ-VLM achieves state-of-the-art accuracy, while attaining competitive performance on 3D VQA and 3D dense captioning datasets compared to methods relying on explicit 3D inputs.

Our main contributions are summarized as follows:

\begin{itemize}
\item We propose Occ-VLM, a 3D vision-language model that operates purely on RGB images. Despite relying solely on a single 2D vision encoder, Occ-VLM achieves comprehensive 3D scene understanding, unifying accurate geometric perception and vision-language spatial reasoning.

\item We design the Occ adapter to bridge 2D semantics and 3D spatial reasoning. It effectively lifts the rich semantic knowledge from 2D pre-training into 3D space while preserving the VLM's instruction-following and vision-language reasoning capabilities.

\item Occ-VLM attains state-of-the-art performance on the multi-view 3D semantic occupancy prediction task, demonstrating the effectiveness of our bidirectional 2D-3D modeling.

\item Extensive experiments on 3D VQA and 3D dense captioning benchmarks reveal that Occ-VLM achieves leading performance among 2D-input 3D VLMs and performs competitively against models that rely on explicit 3D inputs.
\end{itemize}

\section{Related Work}

\subsection{2D Vision Language Models}

Vision-Language Models (VLMs) have achieved significant progress in joint image-text understanding. Existing 2D VLMs typically employ vision-language pretrained Vision Transformers (ViTs) as vision encoders to extract semantically discriminative features. A key design choice lies in the multimodal projector that bridges visual representations and the token embedding space of language models. One line of work, represented by BLIP-2 \cite{li2023blip}, InstructBLIP \cite{dai2023instructblipgeneralpurposevisionlanguagemodels}, Flamingo \cite{alayrac2022flamingo}, and Qwen-VL \cite{bai2023qwen}, adopts the Q-Former architecture that employs learnable queries to aggregate visual information. An alternative approach, exemplified by LLaVA \cite{liu2023visual} and its variants \cite{liu2024improved, li2024mini, wang2024qwen2, lin2024video, liu2024llava}, utilizes lightweight MLP-based projectors to directly map visual tokens into the language model's embedding space. Recent works such as LLaVA-One-Vision \cite{li2024llava, an2025llava} extend these architectures to support both image and video inputs within a unified framework, enabling spatiotemporal reasoning across visual modalities.

\subsection{3D Vision Language Models}
Building upon the success of VLMs in 2D tasks, increasing attention has shifted toward extending their capabilities to 3D domains. Existing 3D VLMs can be categorized into three types based on their input modalities: point-based, RGB-D-based, and RGB-based methods. Point-based methods \cite{guo2023point, qi2024gpt4point, qi2024shapellm, tang2024minigpt, xu2024pointllm} directly adopt point clouds as 3D visual input. These approaches leverage point-cloud encoders pretrained on 3D recognition tasks to extract scene-level or object-level features, which are then projected into compact 3D visual tokens via a Q-Former or MLP. In parallel, RGB-D-based methods \cite{wang2023chat, hong20233d, chen2024ll3da, huang2024chat, fu2024scene, majumdar2024openeqa, man2024lexicon3d, zhu2024llava, zheng2025video, qi2025gpt4scene} take multi-view RGB-D images as input. They typically extract dense visual features from RGB images using pretrained 2D encoders, and then lift these features into 3D space by leveraging the associated depth maps, yielding spatially consistent 3D visual tokens. RGB-based methods have recently emerged as an alternative that relies solely on RGB inputs. These models typically employ a dual-encoder architecture  \cite{wu2025spatial, zheng2025learning, cheng20253d} or neural rendering techniques \cite{petit2025llava, thai2025splattalk} to encode scene geometry. Despite these advances, existing methods either rely on explicit 3D inputs (limiting practicality and scalability) or adopt complex architectures that compromise simplicity and extensibility. To address this, we explore a concise and efficient 3D VLM based solely on RGB inputs, built upon the native VLM architecture.

\subsection{3D Occupancy Prediction}
Occupancy networks \cite{mescheder2019occupancy, peng2020convolutional} were originally proposed as an implicit scene representation to determine the occupancy status of arbitrary 3D coordinates. Specifically, the 3D space is discretized into uniform voxel grids, and models predict occupancy status for each cell given point clouds or images as input. In the context of autonomous driving, subsequent work \cite{cao2022monoscene, huang2023tri, li2023voxformer, huang2024gaussianformer, wei2023surroundocc, pan2024renderocc} augments occupancy grids with semantic labels to enable fine-grained scene understanding. More recently, embodied occupancy prediction \cite{huang2023embodied, wu2025embodiedocc, wang2025embodiedocc++} has emerged as a related paradigm that performs egocentric monocular occupancy estimation and incrementally aggregates predictions into a global representation through continuous agent-scene interaction. This formulation enables globally consistent 3D scene modeling, making it well-suited for embodied AI applications.

\section{Method}

\subsection{Overview}

\begin{figure*}[h]
    \centering
    \includegraphics[width=\textwidth]{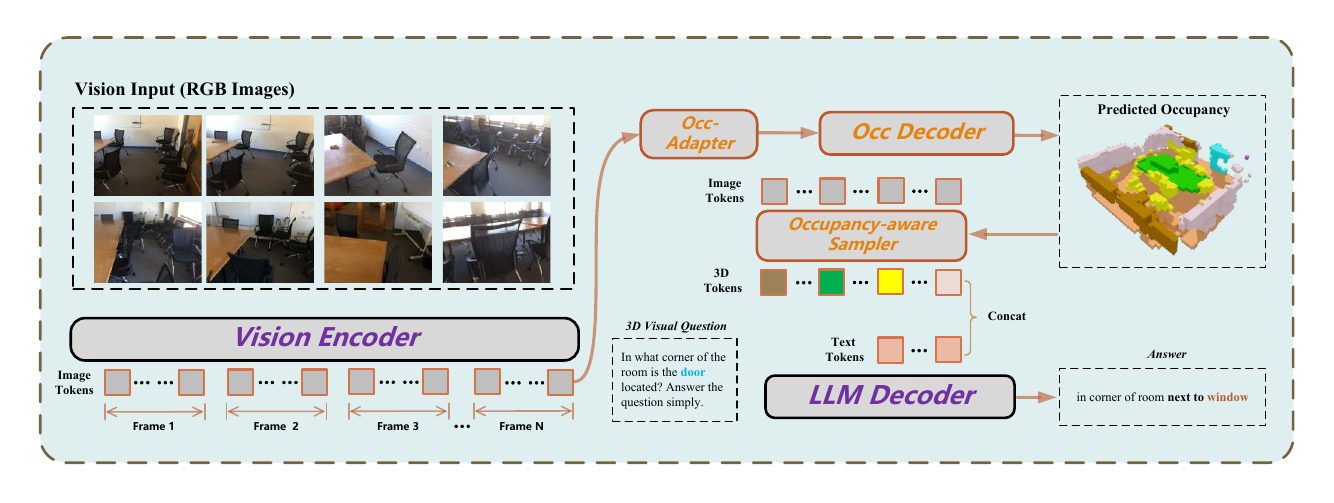}
    \caption{Architecture of Occ-VLM. Taking multi-view RGB images as input, we first extract image tokens using a 2D vision encoder. These tokens are then processed by the Occ Adapter and Occupancy Decoder to predict scene occupancy. Subsequently, we sample foreground tokens guided by the predicted occupancy and incorporate 3D geometric cues through 3D Positional Embeddings. These occupancy-aware tokens are finally input into an LLM decoder for unified vision-language reasoning tasks.}
    \label{fig:framework}
\end{figure*}

In this section, we present \textbf{Occ-VLM}, a vision-language model designed for 3D indoor scene understanding from posed multi-view images. Our approach begins by encoding each input frame into multi-view tokens using a frozen 2D vision encoder. However, these tokens lack explicit 3D structural information. To address this, we introduce a \textbf{Occ adapter} that enables the model to estimate 3D scene occupancy while keeping the 2D vision encoder parameters fixed (Sec. 3.2). Leveraging the predicted occupancy, we construct an \textbf{Occupancy-aware 3D Token Representation} and perform structured sampling of 3D tokens in foreground regions to concentrate on geometrically relevant areas (Sec. 3.3). Finally, we propose a two-stage training strategy that jointly facilitates occupancy reasoning and empowers the model to generate semantically coherent responses for 3D scene question answering (Sec. 3.4). Fig. \ref{fig:framework} illustrates the architecture of our proposed Occ-VLM.

\subsection{Adapting 2D Vision Encoder to Predict Scene Occupancy}

\begin{figure*}[t]
    \centering
    \includegraphics[width=\textwidth]{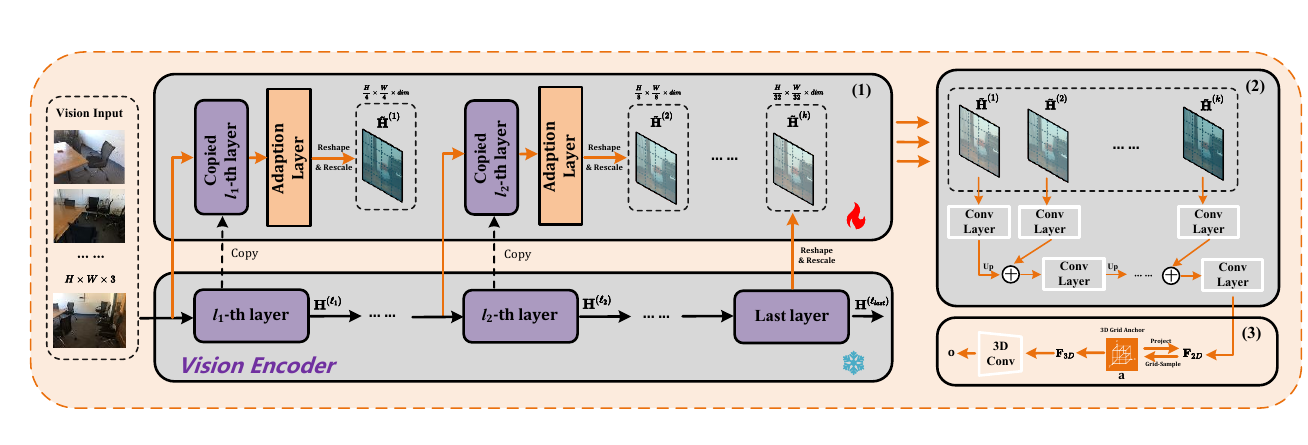}
    \caption{Architecture of the proposed Occ adapter. The adapter consists of three key components: (1) Mid-layer Activation, which selectively extracts multi-scale semantic features from intermediate layers of the frozen vision encoder; (2) Multi-level Token Aggregation, which hierarchically fuses these features to integrate coarse-to-fine representations. (3) 3D Occupancy Decoding, which samples 2D features via anchor projection and grid sampling, and employs 3D convolutions to predict 3D occupancy.}
    \label{fig:occ_adapter}
\end{figure*}

To more effectively capture the geometric attributes of multi-view tokens within 3D LVLMs, we introduce the \textbf{Occ adapter} (Fig. \ref{fig:occ_adapter}). This module equips the model with dense geometric perception capabilities while fully preserving its pre-trained vision-language proficiency. Based on the predicted 3D occupancy, we spatially associate multi-view tokens with occupancy grids, seamlessly disentangling foreground tokens (non-empty grids) from background regions.

\noindent \textbf{Mid-layer Activation.} Since the 2D vision encoder relies on massive parameters and large-scale pre-training to achieve robust semantic understanding, fully fine-tuning it for geometric adaptation would inevitably degrade its inherent capabilities and incur substantial computational overhead. Instead, to preserve the original pre-trained parameters, we keep the entire encoder frozen and instantiate trainable copies of a selected subset of intermediate layers $\{\ell_k\}_{k=1}^{K}$. For each of these copied layers, a lightweight Transformer-based adapter $\mathcal{A}_k$ is appended to reduce token dimension via intra-frame self-attention:
\begin{equation}
\begin{gathered}
\tilde{\mathbf{H}}^{(k)} = \mathcal{A}_k\big(\mathbf{H}^{(\ell_k)}\big) = \mathrm{LN}\Big( \mathbf{H}' + \mathrm{FFN}\big(\mathrm{LN}(\mathbf{H}')\big) \Big), \\ 
\mathbf{H}' = \mathrm{LN}\big(\mathbf{H}^{(\ell_k)} + \mathrm{MHSA}(\mathbf{H}^{(\ell_k)})\big),
\end{gathered}
\end{equation}
\noindent where $\mathbf{H}^{(\ell_k)}$ denotes the hidden states from layer $\ell_k$, $\tilde{\mathbf{H}}^{(k)}$ represents the adapted features for downstream 3D occupancy prediction, and $\mathrm{MHSA}$, $\mathrm{FFN}$, and $\mathrm{LN}$ denote Multi-Head Self-Attention, Feed-Forward Network, and Layer Normalization, respectively.

\noindent \textbf{Multi-level Token Aggregation.} We then fuse the adapted tokens across scales utilizing a top-down convolutional feature pyramid. Starting from the deepest layer (coarsest resolution), features are progressively upsampled and aggregated with finer-level tokens:
\begin{equation}
\mathbf{F}_{2D}^{(k)} = \mathcal{N}_k(\tilde{\mathbf{H}}^{(k)}) = \mathrm{Conv}\Big( \mathrm{Conv}\big(\tilde{\mathbf{H}}^{(k)}\big) + \mathrm{Upsample}\big(\mathbf{F}_{2D}^{(k+1)}\big) \Big),
\end{equation}
\noindent where $\mathrm{Conv}$ denotes a 2D convolutional layer, $\mathrm{Upsample}$ matches spatial resolutions, and $\mathbf{F}_{2D}^{(k)}$ is the aggregated multi-scale feature map at level $k$. We denote the highest-resolution aggregated feature map as $\mathbf{F}_{2D}$ for subsequent projection.

\noindent \textbf{3D Occupancy Decoding.} In an indoor scene, we process $N$ sampled images, each associated with camera intrinsics $\mathbf{K}_v \in \mathbb{R}^{3\times3}$ and extrinsics $\mathbf{E}_v = [\mathbf{R}_v \mid \mathbf{t}_v] \in \mathbb{R}^{3\times4}$. We back-project a set of 3D grid anchors $\{\mathbf{a}_i\}_{i=1}^{M}$ onto the multi-view features $\mathbf{F}_{2D}$. For each anchor $\mathbf{a}_i$ and view $v$, the 2D sampling coordinate $\mathbf{p}_{i,v}$ is computed via perspective projection:
\begin{equation}
\mathbf{p}_{i,v} = \pi\big(\mathbf{K}_v (\mathbf{R}_v \mathbf{a}_i + \mathbf{t}_v)\big),
\end{equation}
Utilizing differentiable grid sampling $\Phi$, we aggregate the multi-view 2D features into a unified 3D volume:
\begin{equation}
\mathbf{F}_{3D} = \left\{ \frac{1}{|\mathcal{V}_i|} \sum_{v \in \mathcal{V}_i} \Phi_{\mathbf{p}_{i,v}}(\mathbf{F}_{2D,v}) \right\}_{i=1}^M \in \mathbb{R}^{M \times D},
\end{equation}
where $\mathcal{V}_i$ denotes the set of valid views where anchor $\mathbf{a}_i$ projects within the image plane, $|\mathcal{V}_i|$ is its cardinality, and $D$ denotes the feature dimension. Subsequently, a 3D CNN decoder $\mathcal{D}_{3D}$ predicts per-voxel semantic labels:
\begin{equation}
\mathbf{O} = \arg\max\big(\mathrm{softmax}(\mathcal{D}_{3D}(\mathbf{F}_{3D}))\big) \in \{0,1,\ldots,N_c-1\}^M,
\end{equation}
where class 0 indicates empty space and classes $\{1,\ldots,N_c-1\}$ denote specific occupied semantics. The centers of the occupied grids are then extracted as: $\mathcal{G} = \{\mathbf{a}_i \mid \mathbf{O}_i > 0\}$, which serves as the explicit structural prior for subsequent 3D token sampling.

\subsection{Occupancy-aware 3D Token Representation}
\label{sec:token_merging}

Building upon the Occ adapter, we reconstruct multi-view posed images into a 3D semantic occupancy representation, enabling dense scene geometry perception within the 3D VLM framework. Subsequently, exploiting the sparse distributional properties of objects in 3D space, we perform spatially-aware sampling over the multi-view image tokens to effectively align them with the underlying 3D structure.

\noindent \textbf{Occupancy-aware Foreground Token Sampling.}
Given the set of occupied grid centers $\mathbf{G}=\{\mathbf{a}_j\}_{j=1}^{|\mathbf{G}|}$ predicted by the Occ adapter, we extract high-level semantic tokens from the final layer of the 2D encoder, denoted as $\mathbf{H}^{(\ell_{\text{last}})}$. To fully leverage the explicit 3D geometric prior estimated by the Occ adapter and effectively bridge the 3D scene with 2D images, we reuse the spatial projection correspondence established in Eq.~(3). We treat the predicted occupancy grids as a geometric mask. Specifically, for each occupied center $\mathbf{a}_j \in \mathbf{G}$, we project it back onto the image planes of all valid views $v \in \mathcal{V}_j$ and sample the corresponding image features. This process constructs a set of foreground tokens, denoted as $\mathcal{T}_{\text{fg}}$:
\begin{equation}
\mathcal{T}_{\text{fg}} = \Big\{ \mathbf{t}_{j,v} = \Phi_{\mathbf{p}_{j,v}}\big(\mathbf{H}^{(\ell_{\text{last}})}_v\big) \;\Big|\; \mathbf{a}_j \in \mathbf{G}, \, v \in \mathcal{V}_j \Big\}.
\end{equation}
By selectively sampling only the tokens anchored to valid 3D geometric structures, we automatically filter out irrelevant background information. Crucially, since our LLM (Qwen2) is pre-trained on video-language data, we flatten the sampled multi-view spatial tokens along the sequence dimension, treating them analogously to temporal frames in video representations. This formulation allows us to directly leverage the LLM's inherent capacity for processing multi-frame inputs to achieve implicit cross-view interactions, ensuring that the final input sequence compactly captures the holistic consistency of the 3D scene.

\noindent \textbf{3D Position Encoding.} After obtaining the sampled foreground tokens $\mathbf{t}_{j,v} \in \mathcal{T}_{\text{fg}}$, we inject them with occupancy-aware spatial location information. For a token corresponding to the 3D coordinate $\mathbf{a}_j \in \mathbb{R}^3$, we apply sinusoidal positional encoding $\mathrm{PE}(\mathbf{a}_j)$ and add it to the feature:
\begin{equation}
\mathbf{t}^*_{j,v} = \mathbf{t}_{j,v} + \mathrm{PE}(\mathbf{a}_j).
\end{equation}
Since the positional encoding $\mathrm{PE}(\mathbf{a}_j)$ explicitly reflects the 3D structural prior derived from the occupancy prediction, this encoding strategy ensures that the LLM receives tokens that are not only semantically rich but also spatially anchored, facilitating precise 3D reasoning.

\subsection{Training}

We adopt a two-stage training strategy for Occ-VLM. In the first stage, we train the Occ adapter offline. Subsequently, in the second stage, we integrate the pre-trained Occ adapter into the overall Occ-VLM framework to perform visual instruction tuning. The specific details of each stage are elaborated below.

\noindent \textbf{Supervised 3D Semantic Occupancy Prediction.} In the first stage, we train the Occ adapter on a 3D semantic occupancy prediction task, enabling the model to predict volumetric scene occupancy from posed images. We supervise this process using the occupancy GT with the Scene-Class Affinity Loss proposed in MonoScene \cite{cao2022monoscene}, thereby constraining the predictions from both semantic and geometric perspectives. Throughout this stage, the parameters of the 2D encoder are kept frozen. This preserves the vision–language capabilities of the 2D encoder acquired during pre-training, while allowing the Occ adapter to learn an effective mapping from 2D semantics to 3D occupancy grids.

\noindent \textbf{Visual Instruction Tuning.} In the second stage, our objective is to enable the LLM to generate structured and semantically consistent responses from the sampled 3D tokens. Specifically, given the occupancy-aware 3D tokens $\mathcal{T}_{\text{fg}}$ and the textual tokens corresponding to the input prompt, the LLM performs autoregressive decoding. We optimize the model using the standard cross-entropy loss over the predicted token distribution, allowing the LLM to generate correct answers conditioned jointly on the 3D tokens and the linguistic context.

\section{Experiment}

\subsection{Settings}

In this section, we comprehensively evaluate the 3D scene understanding capabilities of Occ-VLM from two perspectives: 3D geometric perception and 3D visual language reasoning capabilities. We begin by introducing the detailed experimental settings.

\noindent \textbf{Benchmarks.} For 3D geometric perception, we evaluate our model on the multi-view 3D semantic occupancy prediction benchmark of EmbodiedScan \cite{huang2023embodied}. The test set of this benchmark contains scenes collected from ScanNet \cite{dai2017scannet}, Matterport3D \cite{chang2017matterport3d}, and 3RScan \cite{wald2019rio}, enabling a thorough assessment of the model’s 3D reconstruction quality and semantic understanding under cross-scene and cross-distribution conditions. Furthermore, we benchmark the model's 3D vision language reasoning capability on SQA3D, ScanQA (for 3D VQA), and Scan2Cap (for 3D dense captioning).

\noindent \textbf{Metrics.} Following standard protocols for the 3D occupancy prediction task, we utilize Intersection over Union (IoU) and mean IoU (mIoU) to quantify prediction accuracy at both category-specific and global levels. For the vision-language tasks, we measure the semantic fidelity and linguistic quality of the generated responses using a comprehensive suite of established metrics: Exact Match@1 (EM@1), BLEU-4 (B-4), METEOR (M), ROUGE-L (R), and CIDEr (C).

\noindent \textbf{Implementation Details.} For the architecture of Occ-VLM, we inherit the model design of LLaVA-Video \cite{zhang2410video}, adopting SigLIP \cite{zhai2023sigmoid} as the 2D vision encoder and Qwen2-7B-Instruct \cite{team2024qwen2} as the LLM decoder. We initialize our model with LLaVA-Video's pre-trained weights and proceed with task-specific fine-tuning. During the first training stage, the Occ adapter is trained for 32 epochs on the EmbodiedScan-Occ dataset to establish the 3D occupancy prediction capability. Subsequently, in the second stage, we conduct visual instruction tuning following the standard LLaVA instruction format, where Occ-VLM is fine-tuned for 3 epochs jointly on the ScanQA, SQA3D, and Scan2Cap datasets.

\begin{table*}[t]
    \centering
    \caption{Quantitative results on the EmbodiedScan multi-view occupancy prediction benchmark. We report performance on \textbf{11} common categories selected from the \textbf{81} categories in the dataset. The best result within each modality is shown in bold.}
    \resizebox{0.95\linewidth}{!}{
    \begin{tabular}{c|c|c|c|cccccccccccc}
        \toprule
        \multirow{2}{*}{Method} & \multirow{2}{*}{Venue} & \multirow{2}{*}{Input} & \multirow{2}{*}{mIoU} 
        & \rotatebox{90}{empty} & \rotatebox{90}{floor} & \rotatebox{90}{wall} & \rotatebox{90}{chair} &  \rotatebox{90}{cabinet}& \rotatebox{90}{door} & \rotatebox{90}{table} & \rotatebox{90}{couch} & \rotatebox{90}{shelf} & \rotatebox{90}{window} & \rotatebox{90}{bed}  \\
        & & & & \textcolor{white}{$\blacksquare$} & \textcolor{orange}{$\blacksquare$} & \textcolor{pink}{$\blacksquare$} & \textcolor{yellow}{$\blacksquare$} & \textcolor{blue}{$\blacksquare$} & \textcolor{cyan}{$\blacksquare$} & \textcolor{green}{$\blacksquare$} & \textcolor{red}{$\blacksquare$} & \textcolor{gray}{$\blacksquare$} & \textcolor{brown}{$\blacksquare$} & \textcolor{purple}{$\blacksquare$}  \\   
        \midrule
            SPVCNN \cite{tang2020searching} &  ECCV'20 & PC & 7.32 & 63.04 & 61.30 & 38.82 & 25.10 & 15.28 & 7.55 & 26.60 & 16.23 & 18.19 & 8.26 & 25.64 \\          
        Cylinder3D \cite{zhu2021cylindrical} & CVPR'21 & PC & 11.52 & \textbf{70.22} & \textbf{63.53} & 44.18 & 41.54 & 20.63 & 17.35 & 34.02 & 35.67 & 26.48 & 15.98 & \textbf{42.89}  \\         
        Mink-Net \cite{choy20194d} &  CVPR'19 & PC & \textbf{15.56} & 69.92 & 60.52& \textbf{51.74} & \textbf{49.44} & \textbf{23.08} & \textbf{24.33} & \textbf{45.77} & \textbf{43.52} & \textbf{29.74} & \textbf{23.02} &39.04\\
        \midrule
        OccNet \cite{tong2023scene} & ICCV'23 & RGB & 8.07 & 37.15 & 46.90 & 25.63 & 20.94 & 13.17 & 18.40 & 26.81 & 22.86 & 13.59 & 13.49 & 26.75 \\
        SurroundOcc \cite{wei2023surroundocc} & ICCV'23 & RGB & 9.10 & 38.54 & 46.17 & 23.55 & 23.04 & 13.60 & 19.15 & 27.79 & 22.28 & 13.11 & 13.72 & 24.32  \\
        EmbodiedScan \cite{huang2023embodied} & CVPR'24 & RGB & 14.52 & 48.81 & 49.57 & 32.32 & 32.12 & 20.95 & 24.30 & 27.95 & 30.51 & 27.30 & 19.03 & 35.40  \\
        SliceOcc \cite{li2025sliceocc} & ICRA'25 & RGB & 15.45 & \textbf{49.32} & \textbf{51.39} & \textbf{33.29} & 31.65 & 22.39 & 25.14 & 29.57 & 32.78 & 26.83 & 19.88 & 38.88 \\
        
        \textbf{Occ-VLM} & & RGB & \textbf{18.41} & 49.03 & 51.10 & 32.31 & \textbf{35.01} & \textbf{24.61} & \textbf{25.86} & \textbf{32.67} & \textbf{37.93} & \textbf{28.51} & \textbf{20.92} & \textbf{39.53}  \\
        
        \bottomrule
    \end{tabular}
    }
    \label{tab:exp_occ}
\end{table*}

\begin{figure*}[t]
    \centering
    \includegraphics[width=0.75\textwidth]{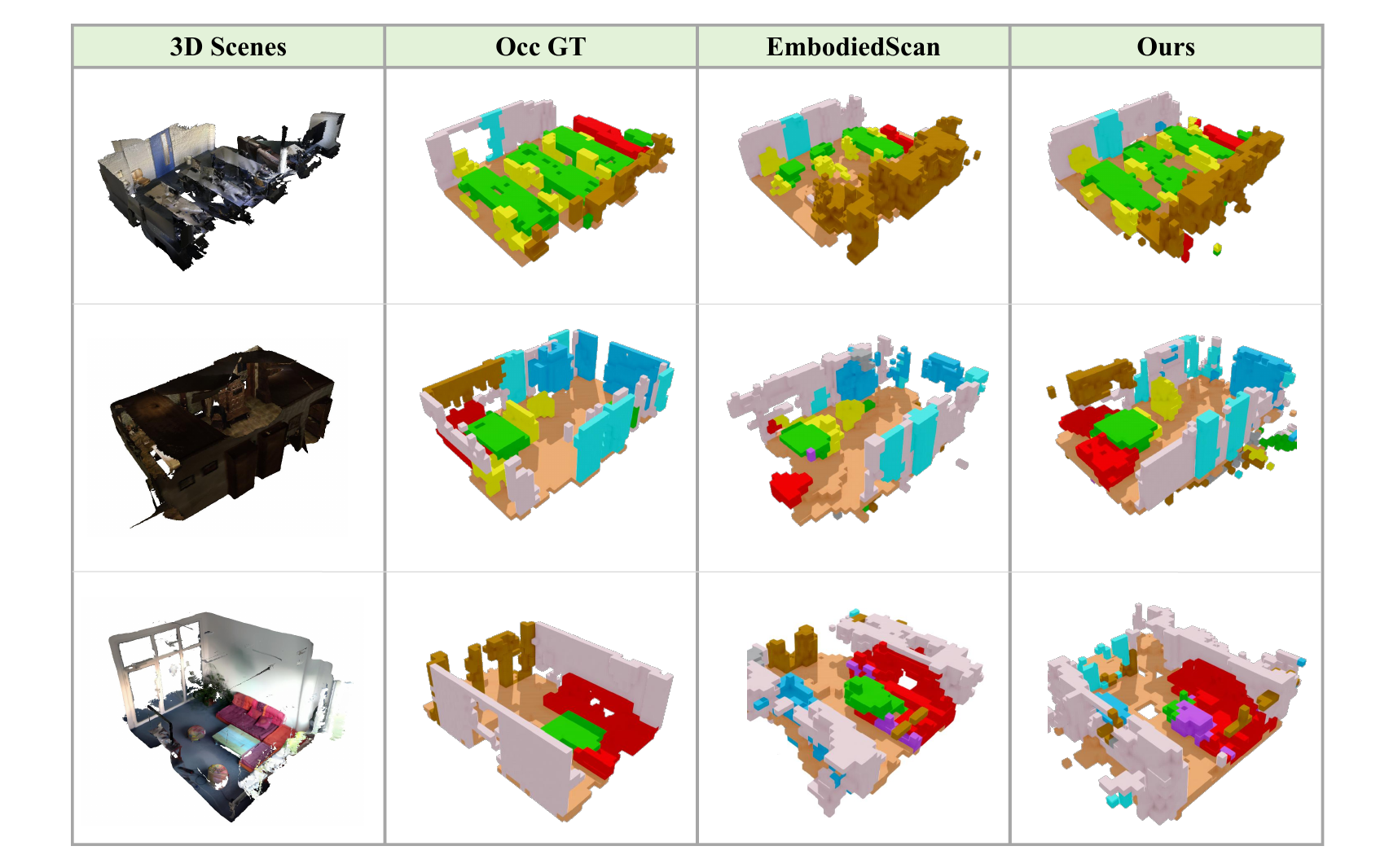}
    \caption{Qualitative results on EmbodiedScan Multi-view 3D semantic occupancy prediction benchmark.}
    \label{fig:vis_occ}
\end{figure*}

\subsection{Evaluation on 3D Occupancy Prediction}

\noindent \textbf{Quantitative results.} We first evaluated the performance of Occ-VLM in 3D geometric perception. Table \ref{tab:exp_occ} presents the results of multi-view 3D semantic occupancy prediction on the EmbodiedScan dataset. To ensure a fair comparison, we selected state-of-the-art occupancy prediction models as baselines, encompassing both point-cloud-based (PC-based) and RGB-based methods. Experimental results demonstrated that Occ-VLM achieved an mIoU of 18.41\% across 81 categories, which not only significantly outperformed all RGB-based methods but also surpassed all PC-based baselines in overall performance. Furthermore, Occ-VLM exhibited substantial gains in prediction accuracy across multiple individual categories. We attributed this performance improvement to the Occ adapter, which bridged 2D semantics and 3D geometry. It successfully translated the rich semantic priors of the pre-trained 2D encoder into spatial cues, revealing a key property: powerful 2D semantic representations inherently facilitated better 3D geometric perception. By lifting these 2D semantic features, the model achieved a significantly more accurate comprehension of 3D scenes.

\noindent \textbf{Qualitative results:} We present the visualizations of Occ-VLM for the 3D semantic occupancy prediction task in Fig. \ref{fig:vis_occ}. Three representative scenes from the ScanNet and 3RScan datasets were selected for qualitative comparison. Specifically, for each scene, the figure sequentially displays the scene mesh, the occupancy GT, the prediction of EmbodiedScan, and the prediction of our Occ-VLM. Qualitative comparisons demonstrated that, compared to EmbodiedScan, Occ-VLM achieved noticeably higher semantic accuracy and a more complete geometric reconstruction.

\subsection{Evaluation on 3D VQA}

\begin{table}[t]
\centering
\caption{Quantitative comparison results on the ScanQA and SQA3D benchmarks. We compared our method with three categories of approaches: (1) task-specific models, (2) 3D-input 3D VLMs, and (3) 2D-Input 3D VLMs. For each method category, we highlighted the best performance metrics in bold.}
\label{tab:3dvqa}
\setlength{\tabcolsep}{5pt}
\resizebox{0.95\linewidth}{!}{%
\begin{tabular}{c|c|c|ccccc|c}
\toprule
\multirow{2}{*}{Method}  & \multirow{2}{*}{Venue}
 & \multirow{2}{*}{Input} & \multicolumn{5}{c}{\textbf{ScanQA} (val)}  & \multicolumn{1}{c}{\textbf{Sqa3D} (test)}
\\
\cmidrule(lr){4-8}   
\cmidrule(lr){9-9} 
 & & & EM@1  & B-4 & R & M & C &  EM@1 \\
\midrule
\multicolumn{9}{l}{\textit{\textbf{Task-specific}}} \\     
ScanQA \cite{azuma2022scanqa} & CVPR'22  & PC  & 21.1 & 10.1 & 33.3  & 13.1 & 64.9 & -\\
SQA3D \cite{ma2022sqa3d} & ICLR'23  & PC & - & - & -  & - & - & 46.6 \\
3D-Vista \cite{azuma2022scanqa}  & ICCV'23  & PC  & 22.4 & 10.4 & 35.7  & 13.9 &  69.6 & 48.5 \\
\midrule
\multicolumn{9}{l}{\textit{\textbf{3D-input VLM}}} \\  
3D-LLM \cite{hong20233d} &  NIPS'23  & PC+RGB  & 20.5 & 12.0 & 35.7 & 14.5  & 69.4 & -\\
LL3DA \cite{chen2024ll3da}  & CVPR'24 & PC  & -  & 13.5 & 37.3 & 15.9 & 76.8 & - \\
Chat-Scene \cite{huang2024chat} &  NIPS'24   & PC & 21.6 & 14.3 &41.6 &18.0 &87.7 & 54.6 \\
Scene-LLM \cite{huang2023embodied}  & WACV'25  & PC+RGB & 27.2 & 12.0 & 40.0 & 16.6 & 80.0 & 54.2 \\
Video-3D-LLM \cite{zheng2025video} & CVPR'25  & RGBD & 30.1 & 15.8 & 49.0 & 19.8   & 102.1 & 58.5 \\
LLaVA-3D \cite{zhu2024llava} & ICCV'25  & RGBD & 30.6  & 16.4 & 49.6 & 20.8   & 103.1 & 60.1\\
\midrule
\multicolumn{9}{l}{\textit{\textbf{2D-input VLM}}} \\  
LLaVA-OV \cite{xing2025conical} & TMLR'24  & RGB & 17.1    & -   & 29.4 &  13.0  & 50.0 & 25.4\\
LLaVA-Video & TMLR'25  & RGB & -  & 3.1 &  44.6  & 17.7 & 88.7 & 48.5 \\
SplatTalk \cite{thai2025splattalk} & ICCV'25 & RGB & 22.4  & -  & 38.5 & 15.6 & 77.5 & 49.4 \\
Spatial-MLLM \cite{wu2025spatial} & NIPS'25  & RGB & 26.3  & 14.8  & 45.0 & 18.4  & 91.8 & 55.9 \\
LLaVA$^3$ \cite{petit2025llava} & AAAI'26 & RGB & 26.0  & -  & 39.6 & 15.8 & 77.6 & - \\
Occ-VLM (Ours) &   & RGB & \textbf{29.6} & 15.7 & 48.2 & 19.5 & 100.0 & \textbf{58.5} \\
\bottomrule
\end{tabular}%
}
\end{table}

\begin{figure*}[ht]
    \centering
    \includegraphics[width=0.95\textwidth]{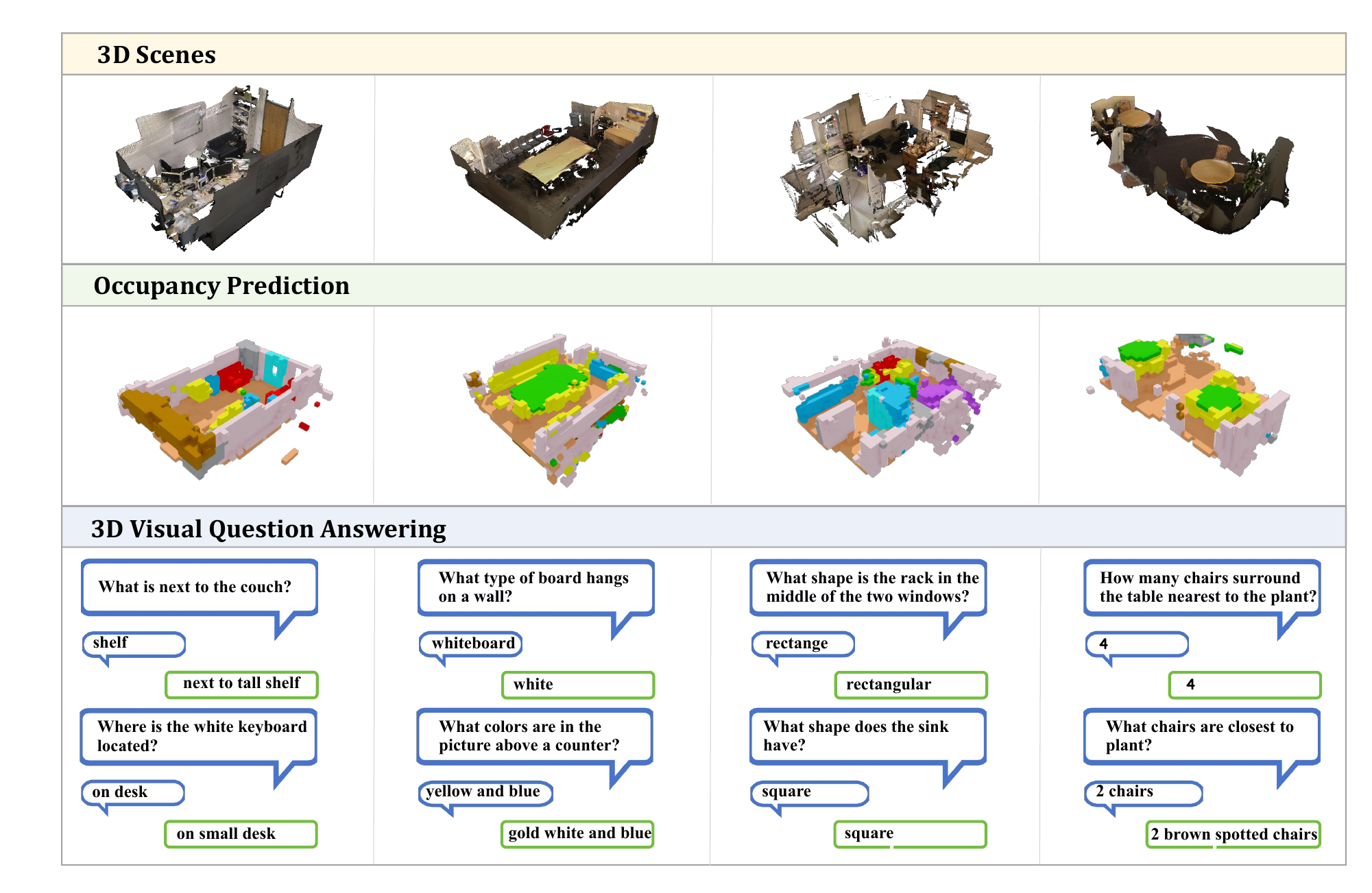}
    \caption{Qualitative results of Occ-VLM on the ScanQA validation set, showing 3D occupancy prediction and 3D VQA outputs across four scenes.}
    \label{fig:vis_qa}
\end{figure*}

We evaluated Occ-VLM on the 3D VQA task, with experimental results on the ScanQA validation set and the SQA3D test set reported in Table \ref{tab:3dvqa}. To facilitate a clear comparison, we categorized the baselines into three groups: (1) \textbf{task-specific models} designed exclusively for 3D VQA; (2) \textbf{3D-input VLMs} that ingested point clouds or RGB-D data; and (3) \textbf{2D-input VLMs} that relied solely on RGB images, encompassing both dual-encoder and neural-rendering-based approaches. As Occ-VLM utilized only RGB inputs during both training and inference, it fell into the 2D-input VLM category.

\noindent \textbf{Quantitative results.} On the ScanQA validation set, Occ-VLM achieved an EM@1 score of 29.6\%, outperforming the 2D-input SOTA Spatial-MLLM. Notably, while Spatial-MLLM adopts a dual-encoder architecture, our model relies on a single 2D vision encoder and attains higher accuracy with a more lightweight design. Compared to neural-rendering-based methods such as SplatTalk and LLaVA$^3$, Occ-VLM demonstrated a clear advantage, improving the EM@1 score by 3.9\%. This margin widened further when compared to video-based baselines (e.g., LLaVA-OV and LLaVA-Video), emphasizing the effectiveness of our occupancy-aware 3D token representation. Furthermore, Occ-VLM remained highly competitive against 3D-input VLMs, surpassing the point-cloud-based LL3DA and 3D-LLM by 4.7\% and 5.8\% EM@1, respectively. Notably, the performance gap between Occ-VLM and the state-of-the-art 3D-input method was a marginal 1.0\%.

Occ-VLM also demonstrated robust performance on the SQA3D test set. Compared to task-specific models such as 3D-Vista, it improved the EM@1 score by a substantial 10\%. Among 2D-input VLMs, Occ-VLM surpassed Spatial-MLLM by 2.6\% EM@1. Moreover, relative to 3D-input VLMs, it outperformed Chat-Scene and Scene-LLM by 1.7\% and 2.1\% EM@1, respectively, and achieved comparable performance to the strongest 3D-input baselines (e.g., Video-3D-LLM and LLaVA-3D). Overall, these results indicate that despite relying entirely on 2D image inputs, Occ-VLM is highly competitive with models that leverage explicit 3D information for visual reasoning, highlighting its robust cross-modal understanding and superior spatial semantic modeling capabilities.

\noindent \textbf{Qualitative results.} In Fig. \ref{fig:vis_qa}, we visualize the performance of Occ-VLM across four ScanNet scenes. The three rows display the scene mesh, the occupancy prediction generated by Occ-VLM, and the corresponding 3D VQA results, respectively. The four columns represent different scenes where we highlight specific capabilities, including spatial relationship reasoning, object recognition, attribute identification (e.g., color, material, and shape), and object counting. As evidenced by these visualizations, Occ-VLM demonstrates the capability to simultaneously perform reliable scene structure prediction and vision-language reasoning, thereby exhibiting a comprehensive understanding of 3D scenes.

\begin{table}[t]
\centering
\caption{Quantitative comparison results on the Scan2Cap benchmark. We compared our method with three categories of approaches: (1) task-specific models, (2) 3D-input 3D VLMs, and (3) 2D-Input 3D VLMs. For each method category, we highlighted the best performance metrics in bold.}
\label{tab:scan2cap}
\setlength{\tabcolsep}{6pt}
\scalebox{0.82}{
\begin{tabular}{c|c|c|cccc}
\toprule
\multirow{1}{*}{Method}  & \multirow{1}{*}{Venue}
 & \multirow{1}{*}{Input} & C@0.5  & B-4@0.5 & M@0.5 & R@0.5 
 % \multicolumn{4}{c}{\textbf{ScanQA} (val)}  
\\

\midrule
 \multicolumn{7}{l}{\textit{\textbf{Task-specific}}} \\     
Scan2Cap \cite{chen2021scan2cap} & CVPR'21 & PC  & 39.1 & 23.3 & 22.0  & 44.8 \\
Vote2Cap-DETR  \cite{chen2023end} & CVPR'23 & PC  & \textbf{61.8} & \textbf{34.5} & \textbf{26.2}  & \textbf{54.4} \\
\midrule
 \multicolumn{7}{l}{\textit{\textbf{3D-input VLM}}} \\  
LL3DA \cite{chen2024ll3da}  & CVPR'24 & PC  & 65.2  & 36.8 & 26.0 & 55.0  \\
Chat-Scene \cite{huang2024chat} &  NIPS'24   & PC & 77.1 & 36.5 & - & -  \\
Video-3D-LLM \cite{zheng2025video} & CVPR'25  & RGBD & \textbf{80.0} & 40.2 & 28.5 & 61.7  \\
LLaVA-3D \cite{zhu2024llava} & ICCV'25  & RGBD & 79.2  & \textbf{41.1} & \textbf{30.2} & \textbf{63.4}   \\
\midrule
 \multicolumn{7}{l}{\textit{\textbf{2D-input VLM}}} \\  
Occ-VLM (Ours) &   & RGB & \textbf{76.5}  & \textbf{38.9} & \textbf{28.2} & \textbf{60.8} \\
\bottomrule
\end{tabular}%
}
\end{table}

\subsection{Evaluation on 3D Dense Caption}

We further evaluated Occ-VLM on the 3D dense captioning task. Unlike 3D VQA, this task requires the model to generate detailed descriptions for specific objects marked by 3D bounding boxes. This places higher demands on the model's ability to understand fine-grained scene details. We evaluated our approach on the Scan2Cap dataset, comparing it with task-specific models and 3D-input VLMs.

\noindent \textbf{Quantitative results.} As shown in Table \ref{tab:scan2cap}, Occ-VLM clearly outperformed 3D-input task-specific models. Furthermore, it remained highly competitive against 3D-input VLMs. Specifically, Occ-VLM surpassed the point-cloud-based LL3DA across all standard metrics. It also achieved performance comparable to the state-of-the-art Video-3D-LLM and LLaVA-3D, with only a marginal gap of 0.3\% and 2.0\% in the ROUGE@0.5 metric, respectively.

\subsection{Ablation study}

\noindent \textbf{Ablation for token representation.} To evaluate the effectiveness of the proposed occupancy-aware 3D token representation, we conducted an ablation study, as detailed in Table \ref{tab:ab1}. These experiments isolated the impact of different token representation strategies while keeping the overall Occ-VLM architecture constant. Specifically, we compared the "Video" setting (where foreground video tokens sampled via occupancy projection are organized in temporal order) with the "Occ Grid" setting (where tokens are spatially aggregated and indexed by 3D occupancy grid centers). Furthermore, we investigated the necessity of explicit spatial information by adding or removing 3D positional encoding (PE). The quantitative results demonstrate that the "Video" representation consistently outperforms the "Occ Grid" approach on both the 3D VQA and 3D dense captioning tasks. This suggests that preserving the fine-grained temporal semantics of video sequences is highly advantageous for complex scene understanding. Crucially, removing 3D PE leads to a severe performance drop across both representation strategies, particularly in 3D dense captioning. This significant decline highlights the essential role of explicit 3D spatial awareness provided by PE, which serves as a vital geometric anchor for precise 3d vision language reasoning and caption generation.

\begin{table}[t]
\centering
\caption{Ablation study on token representation across ScanQA, SQA3D, and Scan2Cap benchmarks.}
\label{tab:ab1}
\setlength{\tabcolsep}{6pt}
\scalebox{0.85}{
    \begin{tabular}{c|cc|c|cc}
    \toprule
    \multirow{2}{*}{\shortstack{\textbf{Token} \\ \textbf{Representation}}}
    & \multicolumn{2}{c|}{\textbf{ScanQA}} 
    & \multicolumn{1}{c|}{\textbf{Sqa3D}} 
    & \multicolumn{2}{c}{\textbf{Scan2Cap}} 
    \\
     & EM@1 & C
     & EM@1 & C@0.5 & R@0.5
     \\
    \midrule
    \text{Video w/o PE} & 29.7 & 100.5 & 57.8 & 30.1 & 58.1 \\
    \text{Video w/ PE} & 29.6 & 100.0 & 58.5 & 76.5 & 60.8 \\
    \midrule
    \text{Occ Grid w/o PE} & 25.4 & 88.2 & 57.6 & 28.6 & 57.3  \\
    \text{Occ Grid w/ PE} & 26.3 & 90.1 & 57.8 & 67.0 & 60.8 \\
    \bottomrule
    \end{tabular}
    }
\end{table}

\begin{table}[t]
\centering
\caption{Impact of varying the number of input views on the model's scene understanding capabilities. The default view number used in the main experiments is marked with $\dagger$.}
\label{tab:ab2}
\setlength{\tabcolsep}{2mm}
\resizebox{0.90\linewidth}{!}{
\begin{tabular}{c|cc|cc|c|cc|c} 
    \toprule
   \multirow{2}{*}{\shortstack{\textbf{Input View} \\ \textbf{Number}}}  & \multicolumn{2}{c|}{\textbf{EmbodiedScan-Occ}} & \multicolumn{2}{c|}{\textbf{ScanQA}} & \multicolumn{1}{c}{\textbf{SQA3D}} & \multicolumn{2}{c}{\textbf{Scan2Cap}} & \multirow{2}{*}{\shortstack{Visual Length \\ Avg.} } \\
    &   mIoU & IoU (Empty) & EM@1 & C & EM@1 & C@0.5 & R@0.5 &\\
    \midrule
   8    & 16.48 & 45.24 & 27.6 & 92.6 & 55.5 & 62.3 & 60.0 &  $1,462$ tokens \\
   16  & 17.86 & 47.76 & 29.0 & 98.0 & 57.2  & 68.8 & 60.8  &  $2,878$ tokens  \\
   24  & 18.29 & 48.42 & 29.2 & 98.6  & 58.3  & 71.9 &  61.0  &  $4,253$ tokens \\
   $32^\dagger$  & 18.41 & 49.03 & 29.6 & 100.0 & 58.5  & 76.5 & 60.8 &  $5,336$ tokens\\
    \bottomrule
\end{tabular}
}
\end{table}

\begin{figure}[t]
    \centering
    \begin{subfigure}[b]{0.45\textwidth}
        \centering
        \includegraphics[width=\textwidth]{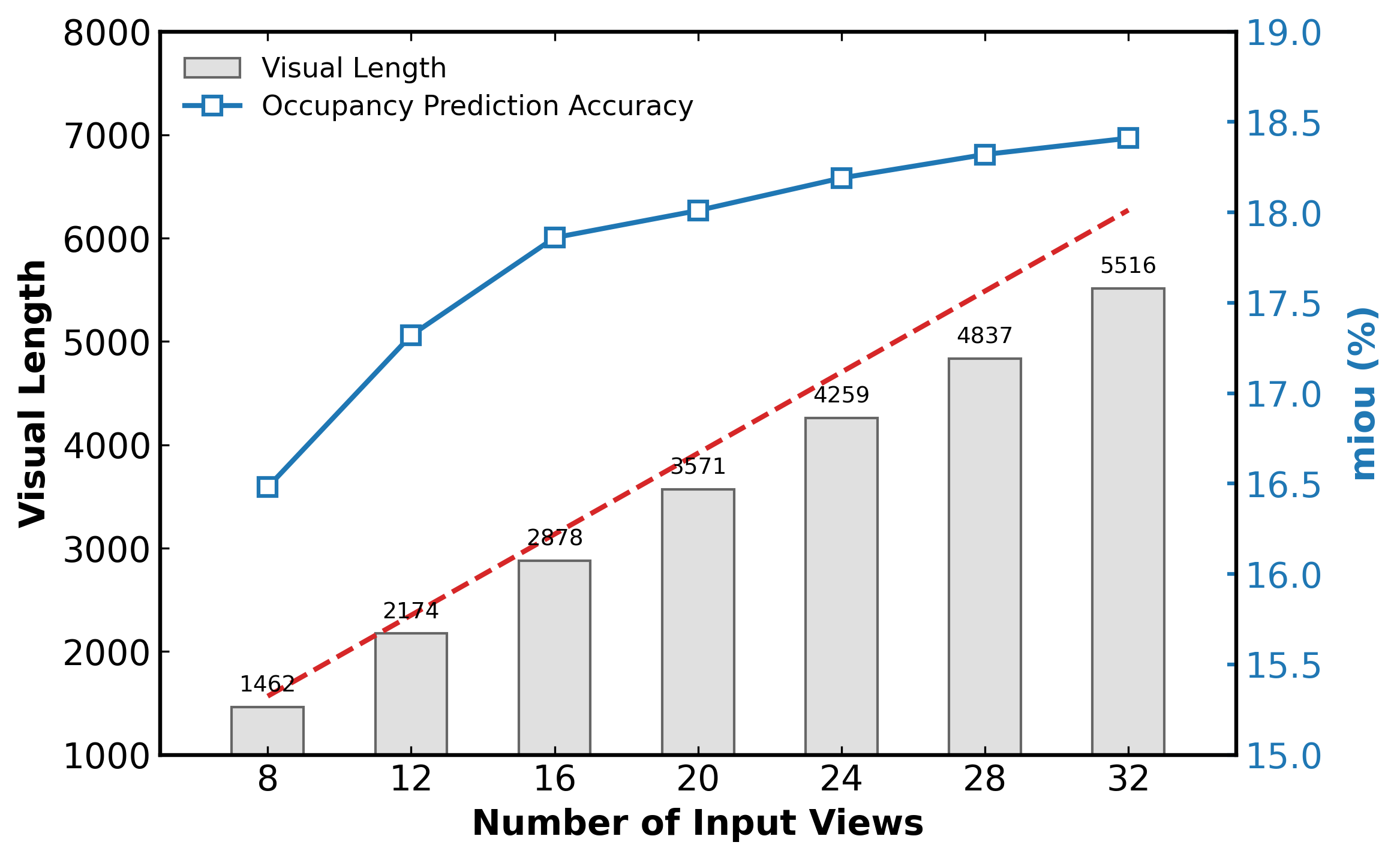} 
        \caption{Occupancy Performance}
        \label{fig:img1}
    \end{subfigure}
    \begin{subfigure}[b]{0.45\textwidth}
        \centering
        \includegraphics[width=\textwidth]{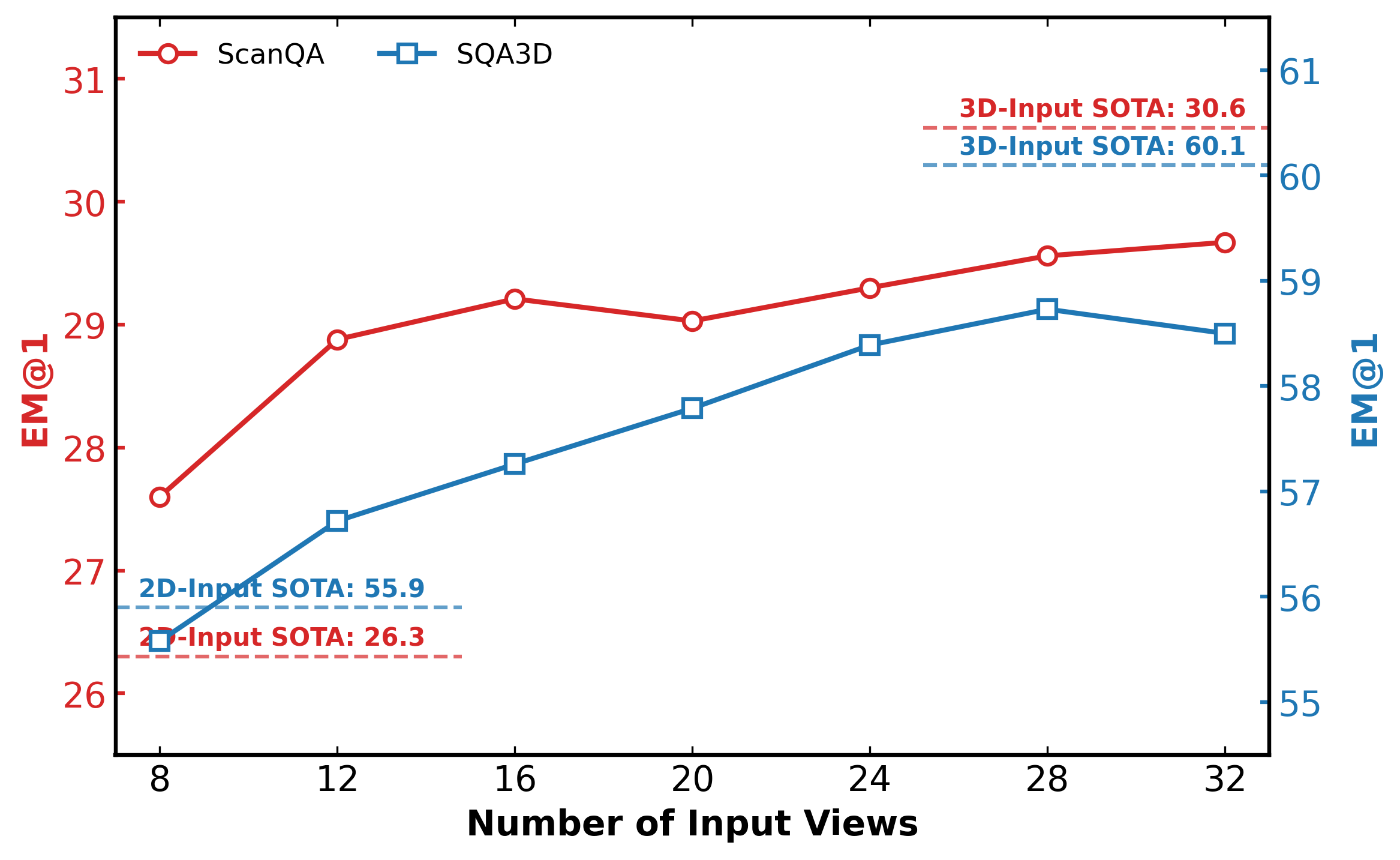}
        \caption{3D VQA Performance}
        \label{fig:img2}
    \end{subfigure}
    \caption{Additional results on the number of input views. More input views lead to higher occupancy prediction accuracy and consistent improvements on 3D VQA and 3D dense captioning.}
    \label{fig:two_images}
\end{figure}

\noindent \textbf{Ablation for input view number.} We investigated how the number of input views affects model performance. As shown in Table \ref{tab:ab2}, the results demonstrate a clear positive correlation between the number of views and both the geometric reconstruction quality and the vision language reasoning capabilities of the model. Specifically, increasing the input views expands the scene's visual coverage and reduces occlusions, which significantly improves the mIoU for occupancy prediction. This improved scene completeness directly benefits downstream vision-language tasks: as the model gathers more comprehensive visual information, performance on ScanQA, SQA3D, and Scan2Cap steadily increases. Fig. \ref{fig:two_images} further illustrates these trends. Notably, as the occupancy prediction mIoU improves, the proportion of foreground tokens relative to total image tokens decreases (see the red line in Fig. 6a). We attribute this to the higher occupancy prediction recall enabled by more input views. This allows the model to localize foreground regions more precisely, thereby effectively filtering out redundant background tokens.

\section{Conclusion}
In this paper, we propose Occ-VLM, a novel 3D vision-language model designed for indoor scene understanding that relies solely on posed RGB images as input. Through the introduced Occ adapter, Occ-VLM effectively utilizes a single frozen 2D vision encoder to simultaneously perform scene geometry perception and 3D vision-language reasoning. Our extensive evaluations reveal a crucial insight: the pre-trained knowledge embedded within a 2D vision encoder can be seamlessly extended to unleash explicit 3D geometric perception abilities. Furthermore, this learned geometry perception serves as a critical spatial guide that fundamentally improves 3D vision-language reasoning. Occ-VLM proves that an RGB-only architecture can achieve profound 3D scene understanding, paving a new direction for this field without the need for explicit 3D inputs.

\newpage
\appendix

% \section{Additional Implementation Details}

% We provide further implementation details regarding the training regimen of Occ-VLM. During the training phase of the Occ-Adapter, we duplicate and activate the 6th, 12th, 18th, and final layers of the 2D vision encoder, optimizing them with AdamW. The base learning rate is set to $1\times10^{-4}$ with a weight decay of 0.05, and the optimizer hyperparameters $\beta$ and $\epsilon$ are set to $(0.9, 0.999)$ and $1\times10^{-8}$, respectively. We apply a module-specific learning rate strategy: the learning rates for the activated encoder layers and their corresponding adaptation layers are scaled by 0.1, whereas the token aggregation module and the 3D occupancy decoder use the base learning rate. The Occ-Adapter is trained for 28 epochs on 8 NVIDIA A800 GPUs with a per-device batch size of 1.

% During the subsequent instruction fine-tuning phase of Occ-VLM, we freeze the vision encoder and only fine-tune the multimodal projection layer and the large language model. The maximum sequence length is set to 12,288, and the model is trained for 2 epochs. Optimization is performed on 4 NVIDIA A800 GPUs with a global batch size of 12, utilizing DeepSpeed ZeRO-3 and gradient checkpointing. We employ a cosine learning rate scheduler with a 3\% linear warmup. The base learning rate is set to $1\times10^{-5}$ (with the vision encoder learning rate specifically set to $2\times10^{-6}$), and weight decay is 0. The entire training pipeline is conducted using bfloat16 mixed precision.

This appendix material contains:
\begin{itemize}
    \item \textbf{Sec. \ref{sec:additional_analysis}} Additional Analysis
    \begin{itemize}
        \item \textbf{Sec. \ref{sec:sqa3d_perf}} Additional Performance on SQA3D
        \item \textbf{Sec. \ref{sec:activated_layers}} Activated Layers of Occ Adapter
        \item \textbf{Sec. \ref{sec:efficiency}} Computational Efficiency and Module Complexity
    \end{itemize}
    \item \textbf{Sec. \ref{sec:qualitative}} Additional Qualitative Results
\end{itemize}

\section{Additional Analysis}
\label{sec:additional_analysis}

\subsection{Additional Performance on Sqa3D}
\label{sec:sqa3d_perf}

\setcounter{table}{5} 
\begin{table}[h]
\centering
\caption{Additional quantitative results on Sqa3D test set. }
\label{tab::ad1}
\setlength{\tabcolsep}{10pt} % 列间距调整为12pt（默认约6pt）
\scalebox{0.8}{
\begin{tabular}{lccccccc}
\toprule
\multirow{2}{*}{Methods} 
 &  \multicolumn{1}{c}{\textbf{EM}} 
 &  \multicolumn{6}{c}{\textbf{Category}} 
 \\
\cmidrule(lr){2-2}
\cmidrule(lr){3-8}   
& EM@1  & What & Is & How & Can & Which  & Others \\
\midrule
 \multicolumn{8}{l}{
\textit{\textbf{Task-specific} methods}
} \\     
% ===== Task-specific Models =====
% ClipBERT \cite{lei2021less}  & 43.3 & 30.2 & 60.1 & 38.7 & 63.3 & 42.5 & 42.7    \\
Sqa3D  & 46.6 & 31.6 & 63.8 & 46.0 & 69.5 & 43.9& 45.3  \\\textbf{
3D-}Vista  & \textbf{48.5} & \textbf{34.8}  & \textbf{63.3}  & \textbf{45.4}  & \textbf{69.8}  & \textbf{47.2}  & \textbf{48.1} \\  
\midrule
 \multicolumn{8}{l}{
\textit{\textbf{3D-input VLM}}
} \\  
% ===== 3D LLM Based Models =====
 
% PQ3D \cite{zhu2024unifying}  &47.1 &  37.1& 61.3 &44.5 &60.9& 47.0 &45.1   \\
Scene-LLM  & 54.2 & 40.9 & 69.1 &45.0 & 70.8 &47.2& 52.3  \\
Chat-Scene  & 54.6  & 45.4 &67.0 & 52.0 &69.5 & 49.9 & 55.0  \\
Video-3D-LLM & 58.4 & \textbf{52.2} & \textbf{72.3} & 54.8 & 69.8 & 50.7 & \textbf{56.0}     \\
\midrule
 \multicolumn{8}{l}{
\textit{\textbf{2D-input VLM}}
} \\  
% ===== Vision LLM Based Models =====
LLaVA-Video   & 48.5  &  42.7 &  56.3  & 47.5  & 55.3 &  50.1  & 47.2 \\ % 98.1
Spatial-MLLM  & 55.9 & 45.9 & 71.6 & 55.1 & 69.5 & \textbf{52.0} & 53.0  \\ % 98.3
Occ-VLM (Ours) & \textbf{58.5}  & 52.0 & 72.2 & \textbf{55.2} & \textbf{71.7} & 51.4 & 55.4    \\ % 99.3
\bottomrule
\end{tabular}}
\end{table}

Table \ref{tab::ad1} reports the additional performance of Occ‑VLM and baseline models across all question types on the SQA3D dataset, along with supplementary analysis. As shown, Occ‑VLM achieved the best results among all 2D‑input VLMs in most question categories, consistent with its overall performance. Moreover, compared to one of the state‑of‑the‑art 3D‑VLMs, Video‑3D‑LLM, Occ‑VLM also surpassed it in certain question types such as “Can” and “Which”.

\subsection{Activated Layers of Occ Adapter}
\label{sec:activated_layers}

\begin{table}[h]
\centering
\caption{Impact of varying the activated layer index of the Occ Adapter. The default setting used in the main experiments is marked with $\dagger$.}
\label{tab:ad2}
\setlength{\tabcolsep}{2mm}
\resizebox{0.90\linewidth}{!}{%
\begin{tabular}{c|cc|cc|c|cc} % 定义列格式：左对齐 | 居中 居中 | 居中 居中 | 居中 居中
    \toprule
   \multirow{2}{*}{\shortstack{\textbf{Activated} \\ \textbf{Layer Index}}}  & \multicolumn{2}{c|}{\textbf{Embodiedscan-Occ}} & \multicolumn{2}{c|}{\textbf{ScanQA}} & \multicolumn{1}{c}{\textbf{SQA3D}} & \multicolumn{2}{c}{\textbf{Scan2Cap}} \\
    &   mIoU & IoU (Empty) & EM@1 & C & EM@1 & C@0.5 & R@0.5 \\
    \midrule
   $\{24\}$    & 16.66 & 46.45 & 28.6 & 96.5 & 58.3 & 68.9 & 60.5  \\
   $\{16, 24\}$  & 17.58 & 47.99 & 29.5 & 99.6 & 58.5  & 73.2 & 61.2   \\
   $\{12 , 16, 24\}$  & 18.30 & 47.82 & 29.6 & 100.1  & 58.6  & 74.1 &  61.3 \\
   $\{8, 12 , 16, 24\}^{\dagger}$  & 18.41 & 49.03 & 29.6 & 100.0 & 58.5  & 76.5 & 60.8 \\
    \bottomrule
\end{tabular}
}
\end{table}

We further analyzed the impact of the activation layer selection for the 2D vision encoder within the Occ Adapter. By default, we activated the $\{8, 12, 16, 24\}$-th layers of the vision encoder, and aggregated the output tokens from these layers for 3D occupancy decoding. In this experiment, we gradually reduced the number of activated layers to observe the performance changes of Occ-VLM in terms of geometric perception and 3D vision-language reasoning. As shown in Table \ref{tab:ad2}, from a geometric perception perspective, reducing the activated layers in the Occ Adapter led to a decline in occupancy prediction accuracy. In particular, when the active layers were reduced from $\{16, 24\}$ to only $\{24\}$, the mIoU decreased by 0.92\%, and the IoU (empty) dropped by 1.54\%. A similar trend was also observed in vision-language reasoning: when the Occ Adapter retained only a single active layer, the model's performance on both 3D VQA and 3D dense captioning tasks dropped significantly. This indicates a strong correlation between the geometric perception capability of Occ-VLM and its vision-language reasoning ability.

\subsection{Computational Efficiency and Module Complexity}
\label{sec:efficiency}

To further clarify the design efficiency of Occ-VLM, we emphasize that our approach avoids the need for explicit depth estimation, point cloud processing, or a separate, heavy 3D geometry encoder. Crucially, the Occ-Adapter shares the 2D visual encoder with the underlying VLM, which minimizes redundant computation. To quantify this module-level efficiency, Table \ref{tab:cost} compares the computational cost and parameter count of our Occ-Adapter against representative 3D geometric encoders (e.g., VGGT and Depth-Anything v2) under a 32-view input setting on ScanNet scenes.

Besides, when computing the cost based only on the activated layers, the Occ-Adapter achieves an ultra-low FLOPs of merely \textbf{2.6T}, compared to dedicated geometry modules (e.g., 38.2T for Depth-Anything v2). This substantiates our design choice of achieving effective 3D reasoning without incurring the heavy overhead of a separate 3D geometry encoder.

\begin{table}[h]
\centering
\caption{Cost comparison of representative geometry modules under a 32-view input setting on ScanNet scenes.}
\label{tab:cost}
\setlength{\tabcolsep}{4mm}
\resizebox{0.82\linewidth}{!}{
\begin{tabular}{c|c|c|c}
\toprule
Module / Model & Input Size & FLOPs ($\downarrow$) & Params ($\downarrow$) \\
\midrule
VGGT & $518 \times 518$ & 124.6T & 1.2B \\
Depth-Anything v2 & $518 \times 518$ & 38.2T & 335.3M \\
\textbf{Occ-Adapter} & $384 \times 384$ & \textbf{11.3T} & \textbf{161.24M} \\
\bottomrule
\end{tabular}}
\end{table}

\section{Additional Qualitative Results}
\label{sec:qualitative}
\setcounter{figure}{6}

\begin{figure}[h]
    \centering
    \includegraphics[width=\linewidth]{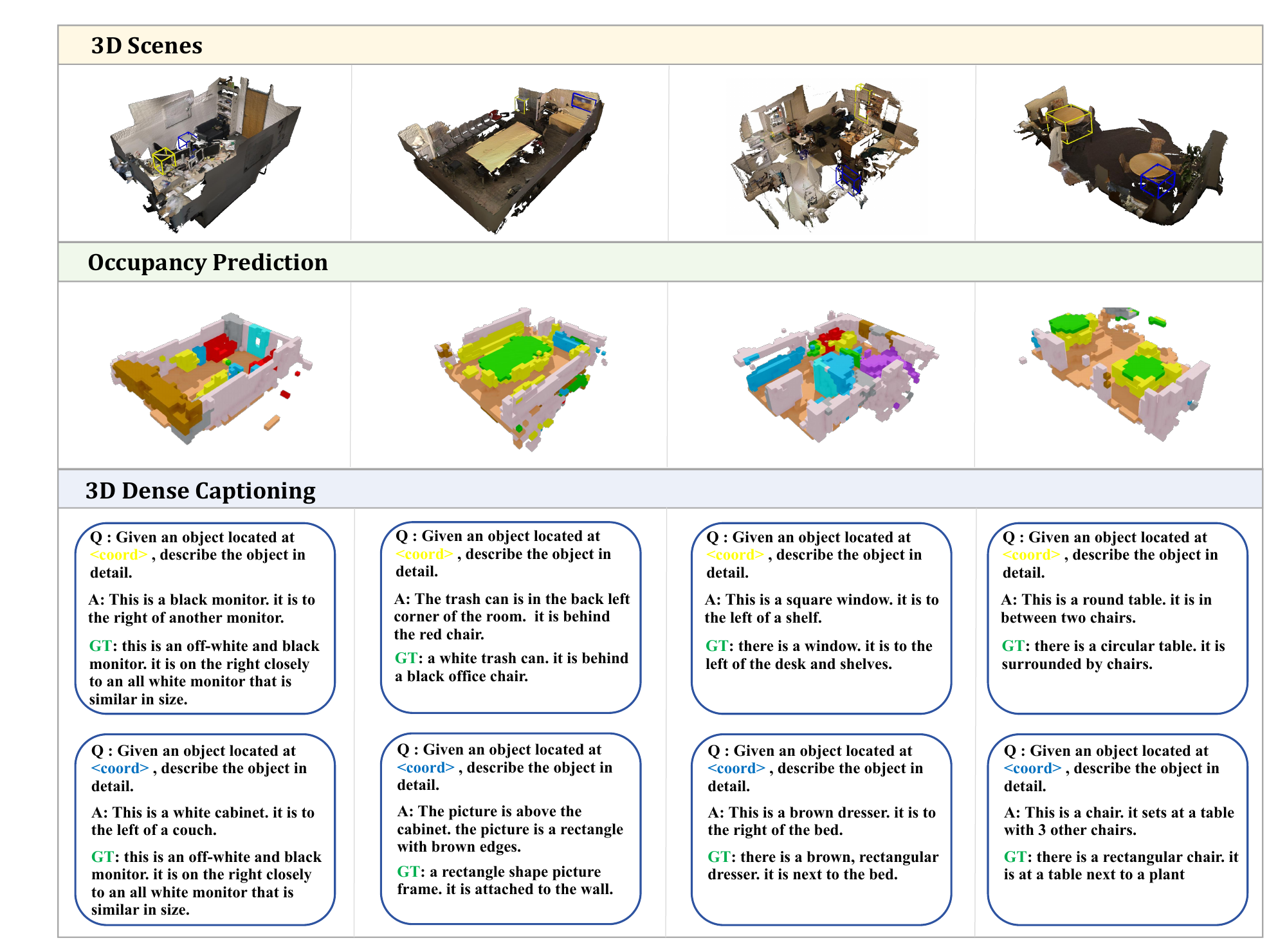}
    \caption{Additional qualitative results on Scan2Cap dataset.}
    \label{fig:viscap}
\end{figure}

We present the visualization results of 3D dense captioning in Fig. \ref{fig:viscap}. This visualization maintains the same scene setup as Fig. 5 in the paper and highlights the corresponding 3D bounding boxes within the context of the caption question.

\bibliographystyle{splncs04}
\bibliography{main}
\end{document}